\documentclass[letterpaper, conference, 10pt]{IEEEconf}
\IEEEoverridecommandlockouts
\usepackage{cite}
\usepackage{amsmath,amssymb,amsfonts}
\usepackage{algorithm}
\usepackage{algpseudocode}
\usepackage{graphicx}
\usepackage{textcomp}
\usepackage{xcolor}
\usepackage{soul}
\usepackage{comment}
\usepackage{bm}
\usepackage{multirow}
\usepackage{dsfont}
\usepackage{ulem}
\usepackage{optidef}
\usepackage{booktabs}

\usepackage[tight]{subfigure}

\newtheorem{Def}{Definition}

\begin{document}

\title{\Large \bf Coupled Active Perception and Manipulation Planning for a  Mobile Manipulator in Precision Agriculture Applications}
\author{Shuangyu Xie,  Chengsong Hu, Di Wang, Joe Johnson, Muthukumar  Bagavathiannan, and Dezhen Song
    \thanks{
    S. Xie, D. Wang, and D. Song are with Department of Computer Science and Engineering, Texas A\&M University. D. Song is also with Department of Robotics, Mohamed Bin Zayed University of Artificial Intelligence (MBZUAI) in Abu Dhabi, UAE. C. Hu, J. Johnson and M. Bagavathiannan are with Department of Soil and Crop Sciences, Texas A\&M University. Corresponding author: Dezhen Song. Email: \texttt{dezhen.song@mbzuai.ac.ae}.
    }
}
\maketitle

\begin{abstract}
A mobile manipulator often finds itself in an application where it needs to take a close-up view before performing a manipulation task. Named this as a coupled active perception and manipulation (CAPM) problem, we model the uncertainty in the perception process and devise a key state/task planning approach that considers reachability conditions as task constraints of both perception and manipulation tasks for the mobile platform. By minimizing the expected energy usage in the body key state planning while satisfying task constraints, our algorithm achieves the best balance between the task success rate and energy usage. We have implemented the algorithm and tested it in both simulation and physical experiments. The results have confirmed that our algorithm has a lower energy consumption compared to a two-stage decoupled approach, while still maintaining a success rate of 100\% for the task.     
\end{abstract}

 % \begin{IEEEkeywords}

 % \end{IEEEkeywords}

\section{Introduction}

In precision agriculture or various mobile manipulation applications, high-resolution scene maps are often inaccessible before the task due to the expensive construction cost or the non-static nature of the scene (e.g., target object moving \cite{qian2022pocd} or growing \cite{4D_crop}). Due to the lack of detailed information on the target object, a typical scenario arises in which the robot first needs to obtain a close-up view of the object of interest before planning the manipulation task. The close-up view provides high-resolution images, facilitating the precise recognition algorithm \cite{Xie_2021}. Based on the precise target information, the mobile manipulator can determine where and how the manipulation should proceed. The process of obtaining the close-up view is an active perception problem. Combining with manipulator task planning, it leads to a coupled active perception and manipulation (CAPM) problem because the probabilistic distribution of the perception result and the end configuration of the robot in active perception affect the subsequent manipulation problem. 

\begin{figure}[hbtp!]
    \centering    \includegraphics[width=3.3in]{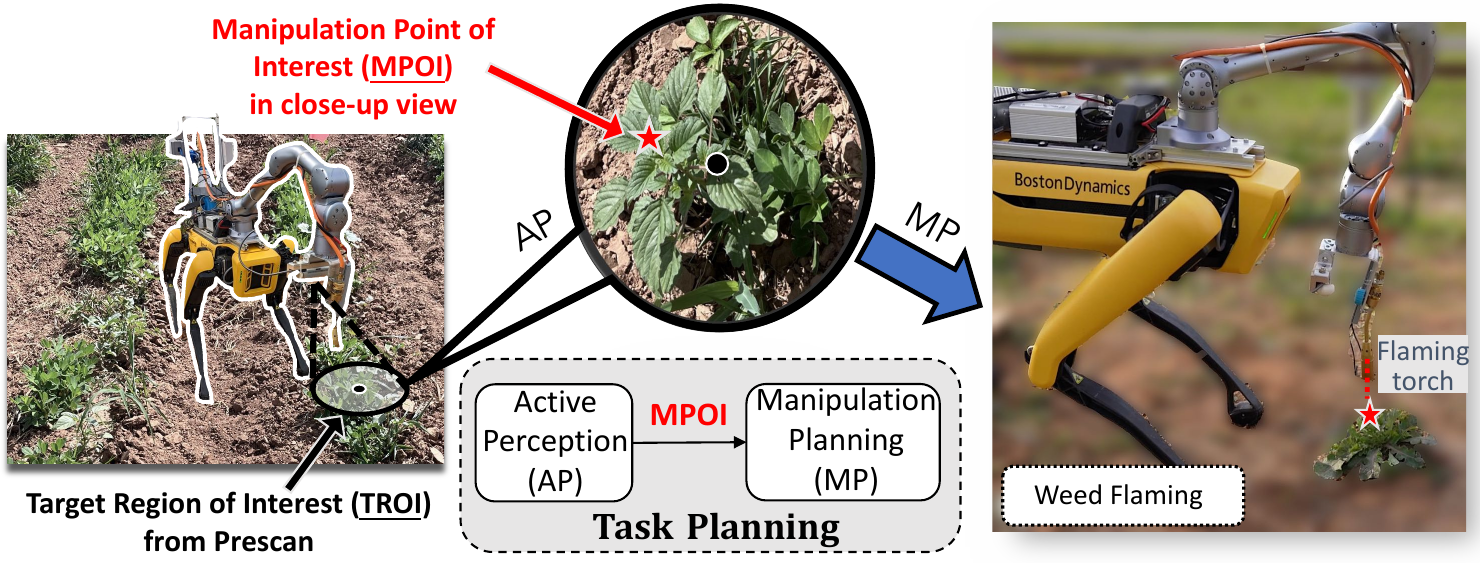}
     \caption{An illustration of CAPM problem scenario using the weed flaming application. }
    \label{fig:nsv}
\end{figure}
Fig.~\ref{fig:nsv} illustrates the CAPM problem using weed removal as an application example where the task is to precisely burn down the biologic center of the weed in the field using the robot. The robot is a hand-eye mobile manipulator with a quadrupedal platform. The robot has a low-resolution prescan of the scene (similar to the background image on the left-hand side in Fig.~\ref{fig:nsv}) to start the task, but the prescan's resolution is insufficient to determine the desired flaming point. With the knowledge of the rough target region of interest (TROI), the robot first navigates towards it to get a close-up view to determine the flaming target position. We name this flaming position as the manipulation point of interest (MPOI), which is shown as the red star in the figure.
Due to the uncertainty of MPOI, the robot has to adjust its next task correspondingly: directly perform the manipulation task (i.e. weed flaming) if the MPOI is within reach or reposition its body for the manipulation. 

The research question is whether this CAPM task planning can be solved in an efficient manner. In this work, we formulate the CAPM problem and analyze four types of problem using reachability analysis by designing the constraints of the active perception and manipulation tasks. We model energy usage in the CAPM problem and propose an algorithm to minimize the expected energy usage by reducing the body reposition motion. We have implemented our algorithm and tested it in both simulation and weed-flaming physical experiments.  The experimental results have confirmed that our algorithm has lower energy consumption compared to a two-stage decoupled approach, while still maintaining a success rate of 100\% for the task.     

% From the task and motion planning perspective, the plan for the task sequence needs to be able to adjust according to the real-time visual feedback so the active perception and task planning need to be solved simultaneously. 
% . 

% \begin{figure}[hbtp!]
%     \centering
%     \includegraphics[page=1,width=3.5in]{Image/Algorithm_illustration.pdf}
%     \caption{ (a) shows a sample of the agriculture scene that needs precision weed management. (b) The low-resolution global map shows the geometric representation of the crop rows and the weed targets. (c) For each target, the active perception and weed removal is considered as the manipulator task planning to generate the feasible end-effector pose and the feasible base placement region. The yellow ellipsoid region is the candidate base placement region. (d) The platform task planning problem is then formulated to find a base state sequence using the candidate placement region. After it reach every target, the manipulator planning will recompute the best pose for actuation. }
%     \label{fig:nsv}
% \end{figure}
\section{Related Work}
% Due to the nature of the CAPM problem, it is closely related to the development of active perception, mobile manipulation, and robotics in precision agriculture. 

% To perform manipulation precisely given the low-resolution prior task information, active perception is considered as the first step to gather the necessary high-resoluted information.
For our task planning, determining the close-up view can be achieved efficiently through active perception. Active perception, by definition, is an intelligent data acquisition process \cite{bajcsy1988active}, which guides the robot to take intentional actions to perceive the required information \cite{bajcsy2018revisiting}. Numerous active perception algorithms are developed for multiple purposes, such as alleviating ambiguity or occlusion issues in object recognition \cite{atanasov2014nonmyopic,ap_obj_class}, improving performance for UAV navigation \cite{ap_rl}, and multi-robot path planning \cite{ap_multi_ro}. The key challenge for active perception is to define the scene-related criteria \cite{bajcsy1988active} as feedback to planning and control (e.g., semantic characteristic \cite{ap_rl}, cross-frame scan overlap \cite{NBV_MM}). For the vision-based sensor, these criteria are achieved by adjusting the viewpoint or the sensor field of view \cite{aloimonos1988active}. However, these works \cite{atanasov2014nonmyopic,ap_obj_class,ap_rl} focus solely on finding the best view without considering the subsequent task. We propose the next sufficient view condition for the active perception constraint as feedback for the robot state. This constraint can be integrated into the planning framework, allowing joint optimization for both active perception and manipulation tasks.

% In our work, we integrate active vision into task planning to improve the overall performance of the system.

% Pioneering research on active vision can trace its roots to the Stanford hand-eye project \cite{feldman1969stanford}. For the hand-eye system, existing work has delved into the development of calibration algorithms \cite{9512030,zhong2023robot}. There is a growing focus on incorporating the planning algorithm with the hand-eye system. Hegedus et al. use the hand-eye system on the mobile manipulator to find the next best view for object observation that guides object grasping \cite{NBV_MM}.  

% However, there is no existing planning framework for the hand-eye system performing the task considering the joint requirement for active perception and manipulation. 

% \textcolor{blue}{(This work, Task-Oriented Active Perception and Planning in Environments with Partially Known Semantics \cite{pmlr-v119-ghasemi20a}, I think it is important but not sure where to go)}

Mobile manipulators have been widely deployed for tasks in factories such as pick-up and delivery \cite{mobile_mani_p&p} or indoor applications such as opening doors \cite{mittal2022articulated}) due to their high dexterity and mobility. 
% In \cite{mittal2022articulated}, the object-centric and agent-centric planning strategy \cite{object_robot_centric} is adopted from manipulator planning to separate kinematic constraints from task space planning in the design of the object-centric interactive plan \cite{interactive_navigation}. 
For long-horizon tasks with sequential nature like chores, the Task and Motion Planning (TAMP) method is designed to discretize the action space into symbolic action with continuous motion. We use a similar formulation by designing the action sequence first and then solving for the state parameters. When scene information is not fully provided, deterministic TAMP approaches face the issue of incorporating it into real-world applications. Recently, there has been progress in the description of the scene that contains uncertainty. Probabilistic modeling of the robot/target state is developed and integrated into the TAMP framework in \cite{adu2021probabilistic,garrett2020online}. These methods have the advantage of dealing with generating sequential action for complicated task space, but they still passively receive the observation instead of actively planning for observation that improve system efficiency.

 To reduce cost and labor dependency, robotic solutions are integrated with precision agriculture applications that include scene perception \cite{Xie_2021, forest_perception, Agr_Stachniss_ICRA_2022, 2022_weed_spray}, robot navigation \cite{agr_navigation-RSS-21}, motion planning \cite{Volkan_Weed_Mowing}, and the deployment of aerial/ground platforms \cite{forest_semantic_slam, Multirobot, agrbot_3.0}. However, due to the nonstatic scene caused by the complex plant morphology, close-up-view observation should be considered with manipulation in task planning.

\section{Problem Formulation}
% Goal 

Our goal is to find an efficient action plan and state parameters for a mobile manipulator to tackle CAPM tasks using weed flaming as an example.
To formulate our problem and focus on the key issues, we have the following assumptions.

\subsection{Assumptions}\label{ssc:assumption}
\begin{enumerate}
\item[a.1] The mobile platform is holonomic in this paper. %The non-holonomic robots will be considered in the future.    
    
\item[a.2] To ensure the stability of the mobile manipulator, we assume that the robot does not execute body and arm motion simultaneously. Consequently, the overall motion plan can be split into disjoint body motion and manipulator motion subplans.    

\item[a.3] We do not consider obstacle avoidance in body motion planning because our weed removal robot stays on top of weeds and crops in agriculture fields. For this reason, the trajectory of body motion is always along the shortest path from the current state to the goal state in each subtask. Therefore, the motion planning problem is reduced to a task state planning problem. 
    
\item[a.4] Since the robot base is significantly heavier than that of the arm, the energy usage of the arm motion is negligible compared to that of the body.

\item[a.5] Camera resolution is fixed.
\end{enumerate}

\begin{comment}
Here we define some common notations:
\begin{itemize}
    \item $\mathbf{x}$ denotes the robot state. 
    \item  $X\in \mathbb{R}^3$ denotes the 3D position.
\end{itemize}
\end{comment}

\subsection{Input, Key States, and Output}

\subsubsection{Initial Input} Prior to the task, the initial state of the robot is given as input. Scene information is provided by a low-resolution field prescan. This prescan can be obtained by drone surveillance or cameras above ground. Let us define the field as the robot workspace $\mathcal{W} \subset \mathbb{R}^3$. With the low-resolution prescan, the object recognition algorithm (e.g.,\cite{li2022yolov6}) can be applied to identify the TROI, which is defined as $R^{\mbox{\tiny TROI}}_w\subset \mathcal{W}$ (as shown in Fig.~\ref{fig:nsv} (a)). Without loss of generality, we model TROI as a half-ball above the groundplane centered on $X_{w}=[x_{w},y_{w},0]^\mathsf{T} \in \mathcal{W}$ and with radius $r_{w}$. It is possible that the workspace contains more than one TROI and the robot has to handle them one at a time. We focus only on the next TROI in this paper, which is defined as,   
    $$R^{\mbox{\tiny TROI}}_w = \{X=[x,y,z]^{\mathsf{T}}:  \|X-X_{w}\|_2^2 \le r_{w}, z\ge 0\} \subset \mathcal{W}.$$
$R^{\mbox{\tiny TROI}}_w$ is a primary input of our problem.

\textbf{Remark:} $X_{w}$ does not necessarily overlap with MPOI $\bar{X}_{w} \in \mathcal{W}$ because $X_{w}$ is the geometrical center position observed from the low-resolution prescan and $\bar{X}_{w}$ is the biological center.
% In the application of weed flaming, $\bar{X}_{w}$ is the best target point for flaming on the weed (as shown in Fig.~\ref{fig:nsv}). 
$\bar{X}_{w}$ cannot be obtained from the prescan data because it requires a close-up view image for recognition~\cite{Xie_2021}. However, we know $\bar{X}_{w} \in R^{\mbox{\tiny TROI}}_w$ from the design of the prescan recognition algorithm. Therefore, $R^{\mbox{\tiny TROI}}_w$ can be used to plan the initial body motion for active perception. In fact, efficiently obtaining $\bar{X}_{w}$ is the active perception part of the problem, which is related to the notion of key states.
    
\begin{figure}[hbtp!]
    \centering
    \includegraphics[page=1,width=3.4in]{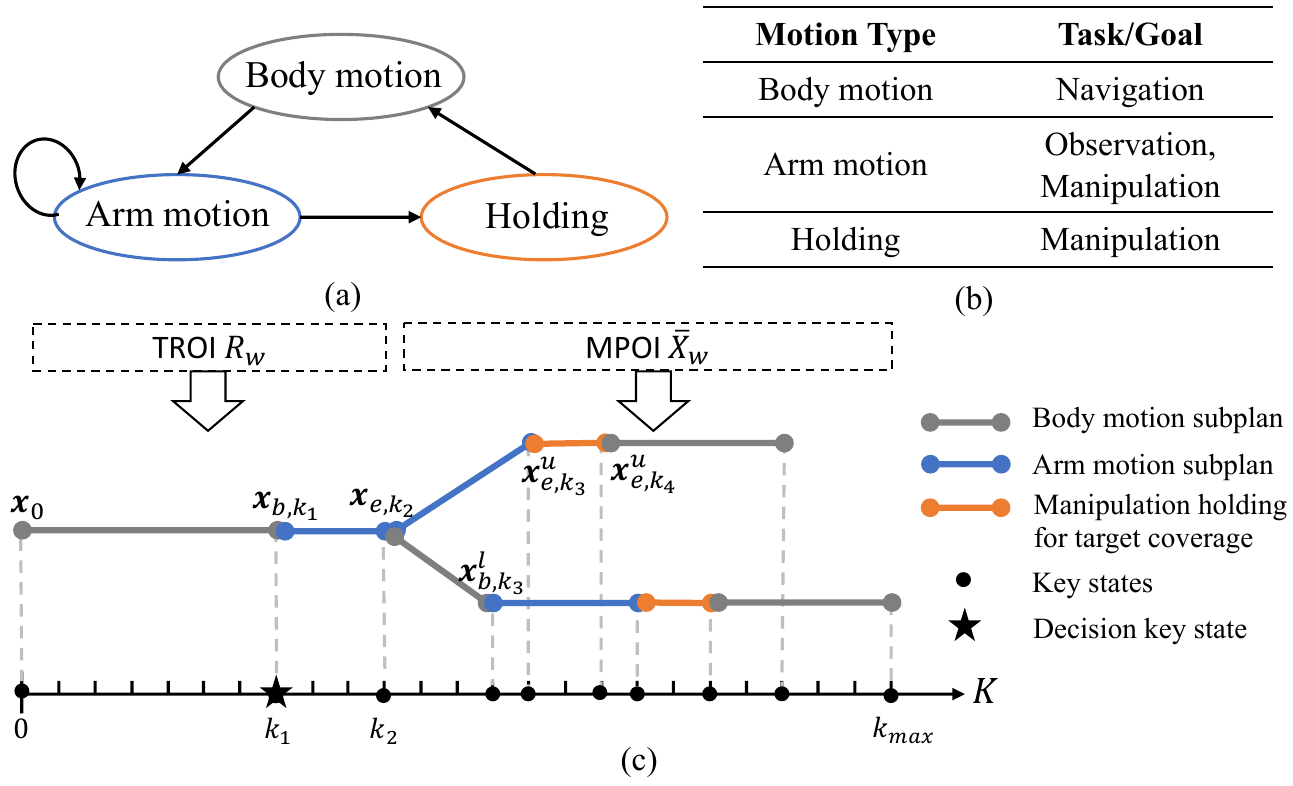}
     \caption{(a) Action transition diagram. (b) Action/task table.  (c) Two possible key state operation sequences.  }
    \label{fig:keystate}
\end{figure}

\subsubsection{Key States} Our state representation includes both the environment states and the robot states. While $R^{\mbox{\tiny TROI}}_w$ and $\bar{X}_{w}$ fully describe the static environment, the robot states are dynamic. Denote the discrete time set $\mathcal{K} = \{0,...,k_{\mbox{\tiny max}}\}$ as the time index set for the entire problem.  Let us represent the state of the robot at time $k$ as $$\mathbf{x}_k:= [\mathbf{x}_{e,k}, \mathbf{x}_{b,k}]^\mathsf{T}.$$ For $k\in \mathcal{K}$, $\mathbf{x}_k$ includes both the pose of the mobile platform (aka body) $ \mathbf{x}_{b,k}\in \mathcal{X}_b\subset SE(3)$ and the manipulator (aka arm) states $\mathbf{x}_{e,k} \in \mathcal{X}_e \subset SE(3)$, where $\mathcal{X}_b$ and $\mathcal{X}_e$ are the state space of body and arm.  Denote the 3D position and orientation of the body in quarternion as $X_{b,k}$ and $\mathbf{q}_{b,k}$. For the arm state, $\mathbf{x}_{e,k}$ is represented by the pose of the end effector rather than by the arm configuration in its joint space.

Key states refer to states acting as decision points that significantly alter the operation of the robot. We consider two types of key state. According to Assumption a.2 in Sect.~\ref{ssc:assumption}, the motion of the robot body and the motion of the arm are executed disjointly. The states that trigger the switching between motion types are one type of key state. Fig.~\ref{fig:keystate}(a) explains the types of robot motion and the switching relationship between them. The other type of key states are the states associated with goal change even with the same motion type. For example, if the arm switches from an active perception task to a manipulation task, the state associated with the switch is also a key state. Fig.~\ref{fig:keystate}(b) explains the correspondence between the type of motion and the goal.
% , and the switching relationship between them.

\subsubsection{CAPM Operation Sequence}\label{ssc:operation-sequence} Fig.~\ref{fig:keystate}(c) illustrates two typical operation sequences in key states. Initially, the robot is in the starting state $\mathbf{x}_0$ and let $k_0 =0$ since the first key state is the starting state. The robot moves its base to $\mathbf{x}_{b,k_1}$, which allows it to move closer to $R^{\mbox{\tiny TROI}}_w$. $k_1$ is the time index when the robot makes its first transition from body motion to arm motion to obtain a close-up observation of TROI. At time $k_2$, the arm reaches $\mathbf{x}_{e,k_2}$ to closely observe $R^{\mbox{\tiny TROI}}_w$ and provides high-resolution images for the recognition algorithm to estimate MPOI $\bar{X}_{w}$.

Depending on $\bar{X}_{w}$ in relation to the pose of the robot, there are two possible subsequent operations depicted as upper and lower branches in Fig.~\ref{fig:keystate}(c). In the upper branch, the robot finds that it can perform task manipulation (i.e. weed flaming) without moving its body. Then it executes the arm motion to reach the flaming pose $\mathbf{x}^u_{e,k_3}$ (the subscript $u$ means upper branch) and stay there until $k_4$ based on the flaming duration requirement before moving to the next target. In the lower branch, since the robot cannot reach $\bar{X}_{w}$, it must be repositioned to $\mathbf{x}^l_{b,k_3}$ (the subscript $l$ means lower branch) before the flaming operation. The following action sequence is the same as that in the upper branch. Since the upper branch and the lower branch have different time lengths, we denote the time index for the ending state as $k_{\mbox{\tiny max}}$, which should also be the last key state. It is clear that the key states are non-deterministic because they are largely dependent on the inherently probabilistic perception result~$\bar{X}_{w}$.

 %The key state $\mathbf{x}_{k_i}$ is a robot state in which the robot base or arm stops and prepares for the next move; e.g. the robot base stops and prepares for observation. Some key states are nondeterministic because they are inaccessible to environmental information. We define the key states as key decision states if their attribute is nondeterministic. The probability that key decision states will satisfy attributes is denoted as $p(\mathbf{x}_{k_i})$. For the rest of the key states, the probability is deterministic.

 \subsubsection{Output}
 Let us define the set of key states as $\{\mathbf{x}_{k_i}| k_i \in \mathcal{K}'\}$, where $\mathcal{K}' \subset \mathcal{K}$ is the set of key state time indexes. In fact, the output can be viewed as a sequence of task planning with each task goal represented as a key state. As stated in Assumption a.3 in Sec.~\ref{ssc:assumption}, our focus is not solving the full motion plan, but the generation of key states. 
It is worth noting that we cannot generate the entire sequence of key states at once with initial input because of the uncertainty in perception. In particular, as the output of active perception, the value of MPOI $\bar{X}_{w}$ observed in the key state of the arm $\mathbf{x}_{e,k_2}$ determines the follow-up key state choices. In addition, $\mathbf{x}_{e,k_2}$ depends on the previous body key state $\mathbf{x}_{b,k_1}$. Later, we will explain those dependencies as active perception and manipulation constraints. For now, this is sufficient for us to introduce the following problem definition.

\subsection{Problem Definitions}
%Our CAPM problem is defined as follows.
\begin{Def}[CAPM Definition]    
 Given the initial state $\mathbf{x}_{k_0}$ and TROI $R^{\mbox{\tiny TROI}}_w$, sequentially plan for key states $\mathbf{x}_{k_1}, ..., \mathbf{x}_{k_{\mbox{\tiny max}}}$, where $k_1,..., k_{\mbox{\tiny max}} \in \mathcal{K}'$, to obtain MPOI $\bar{X}_{w}$ to guide and execute the subsequent manipulation task with the minimum expected energy cost:
\begin{equation}\label{eq:min-exp-energy}
% \mathbb{E}[\sum_{i\in\mathcal{K} } c(\mathbf{x}_{k_{i-1}},\mathbf{x}_{k_{i}})],
\mathbb{E}\left [ \sum_{k_i\in\mathcal{K}' } c(\mathbf{x}_{k_{i-1}},\mathbf{x}_{k_{i}})  \right ]
\end{equation}
where $\mathbb{E}(\cdot)$ is expectation function and $c(\cdot)$ is energy function that will be defined later in the algorithm section. 
\end{Def}

Note that obtaining MPOI $\bar{X}_{w}$ means that certain key states need to satisfy the perception condition. Meanwhile, the completion of the manipulation task means that the task execution condition should be satisfied. Therefore, the definition of the current problem is not complete and \eqref{eq:min-exp-energy} is a constrained optimization problem with conditions introduced in the next section.

\section{Algorithm}

Before we introduce our algorithm, it is important to explain the task constraints associated with active perception and manipulation.

\subsection{Task Constraints}\label{ssc:task-constraints}

\subsubsection{Active Perception Constraints}
To obtain MPOI $\bar{X}_{w}$, the robot must observe TROI $R_{w}$ fully in the field of view and
be close enough so that the image can provide sufficient details.
This leads to view coverage condition and target resolution condition.
These two conditions, which are dubbed the next sufficient view condition as a whole, impose constraints on the pose of the arm $\mathbf{x}_{e,k}$ because the camera is mounted on the end effector. From now on, we drop $k$ from the notation (e.g. use $\mathbf{x}_{e}$ instead of $\mathbf{x}_{e,k}$) for brevity. $k$ can be added back if the temporal index becomes necessary. Denote TROI projection on the ground plane as $R^g_w = \{R^{\mbox{\tiny TROI}}_w | z=0\} $. 

\textbf{View coverage condition}: Keeping the entire $R^g_{w}$ in the camera field of view guarantees that $\bar{X}_{w}$ is visible within the image. Let the camera image resolution be $u_{\mbox{\tiny max}} \times v_{\mbox{\tiny max}}$ pixels.   The entire image can be represented as a set of pixels in a homogeneous coordinate $I := \{\mathbf{p}| \mathbf{p} =[u,v,1]^\mathsf{T},u\in [1,u_{\mbox{\tiny max}}], v \in [1,v_{\mbox{\tiny max}}] \}$. Since the camera image covers the ground plane in which the targets of interest lie, the relationship between the ground plane and the image plane can be characterized as a homography transformation $H_{\mathbf{x}_e}$ from the projective geometry. $H_{\mathbf{x}_e}$ is parameterized by the pose of the end effector $\mathbf{x}_e$. The camera field of view can be back-projected to $\mathcal{W}$ as $R^{\mbox{\tiny FoV}}_w$:
$    R^{\mbox{\tiny FoV}}_w = \{ X 
    | X = (H^{-1}_{\mathbf{x}_e})\mathbf{p},~X \in \mathcal{W},~ \forall\mathbf{p}\in I \}.$
View coverage condition is defined as a binary function:
\begin{equation}
\label{eq:view_coverage}   \mathds{1}_{\mbox{\tiny coverage}}(\mathbf{x}_e,R^{\mbox{\tiny TROI}}_w)=
\begin{cases}
 1,~~ \text{if} ~ R^g_w \subset R^{\mbox{\tiny FoV}}_w  \\ 
0,~~ \mbox{otherwise.} 
\end{cases}
\end{equation}
% Denote the $\{(X_i,Y_i,0)| i = 1,2,3,4\}$ are four corner points be projected to the ground plane can be calculated using \eqref{eq:image2ground}.
% \begin{equation}
% \label{eq:image2ground}
%     \begin{bmatrix}
%     X_1 & X_2& X_3& X_4 \\
%     Y_1 & Y_2& Y_3& Y_4 \\
%     1&1&1&1\\
% \end{bmatrix} = H^{-1}(\mathbf{x}_e) 
% \begin{bmatrix}
%     0 &  u_{\mbox{\tiny max}} & 0 & u_{\mbox{\tiny max}} \\
%     0 & 0& v_{\mbox{\tiny max}}& v_{\mbox{\tiny max}} \\
%     1&1&1&1\\
% \end{bmatrix} 
% \end{equation}

% The line segments corresponding to the corner points can be defined as $\{s_i| i = 1,2,3,4\}$. The condition for view coverage is defined as the through distance function $d$ between line segment and target RoI center:

\textbf{Target resolution condition}: It is important to ensure that the projection area of TROI $R^{g}_w$ into the image coordinate system is large enough so that the object recognition algorithm can effectively detect MPOI $\bar{X}_{w}$. $R^{g}_w$ is projected to an ellipse $R^{\mbox{\tiny TROI}}_I$ in the image coordinate system indicated by the subscript $I$:
$
    R^{\mbox{\tiny TROI}}_I = \{   \mathbf{p}| \mathbf{p}=H_{\mathbf{x}_e}X ,~ \forall X \in R^g_w\}.
$
Let the area of $R^{\mbox{\tiny TROI}}_I$ in the image coordinate system be $\mbox{Area}(R^{\mbox{\tiny TROI}}_I)$. 
Define the area ratio of $R^{\mbox{\tiny TROI}}_I$ in the entire image as the target resolution condition $h(\mathbf{x}_e,R^{\mbox{\tiny TROI}}_w)$:
\begin{align}
\label{eq:target_resolution}
    h(\mathbf{x}_e,R^{\mbox{\tiny TROI}}_w) = \frac{\mbox{Area}(R^{\mbox{\tiny TROI}}_I) }{u_{\mbox{\tiny max}}  v_{\mbox{\tiny max}} } \ge \delta
\end{align}
where $\delta$ is the threshold for the area ratio.

The overall active perception constraint is a combination of view coverage condition and target resolution condition. We name it as the next sufficient view condition (NSV) $\mathds{1}_{\mbox{\tiny NSV}}(\mathbf{x}_e, R^{\mbox{\tiny TROI}}_w)$ as follows,  
\begin{align}
\mathds{1}_{\mbox{\tiny NSV}}&(\mathbf{x}_e, R^{\mbox{\tiny TROI}}_w) = \nonumber\\ \label{eq:observation_constraint}
    &\begin{cases}
 1, & \text{if}~ \mathds{1}_{\mbox{\tiny coverage}} (\mathbf{x}_e,R^{\mbox{\tiny TROI}}_w) ~\bigwedge
 ~\bigl(h(\mathbf{x}_e,R^{\mbox{\tiny TROI}}_w)  \ge \delta\bigr)  \\ 
0,& ~~~~~~~~~~~~~~~~~~ \mbox{otherwise}. 
\end{cases}
\end{align}

\subsubsection{Manipulation Constraints}
To execute the manipulation task, for example weed flaming, it is necessary that the manipulator maintains a certain pose for a given amount of time, which are the spatiotemporal manipulation constraints.  % Denote the robot end effector state at timestamp $k$ as $\mathbf{x}_{e,k}$.

\textbf{End-effector pose condition for manipulation (EPMC)}: This is the condition to determine whether the manipulator can reach the target but not too close according to the task requirement. Being too close may cause self-collision or inability to execute weed flaming task without damaging the robot. For a given MPOI $\bar{X}_{w}$ and the state of the end effector $\mathbf{x}_{e,k}$, we define EPMC as a binary function $\mathds{1}_{\mbox{\tiny EPCM}}(\mathbf{x}_{e,k},\bar{X}_{w})$ at time $k\in\mathcal{K}_m$, where $\mathcal{K}_m \subset \mathcal{K}$ represents a discrete time set during the manipulation period. 
Recall that $\mathbf{x}_{e,k}\in  SE(3)$ is the pose of the end effector, let
$X_{e,k} \in \mathcal{W}$ be its position components, and let the orientation of the end effector point to the MPOI. Define $\varepsilon_{\mbox{\tiny min}}$ and $\varepsilon_{\mbox{\tiny max}}$ as the minimum and maximum distance thresholds for the end effector to be able to perform the task, respectively. EPCM is defined as
\begin{equation}
\label{eq:manipulator_coverage}
\mathds{1}_{\mbox{\tiny EPCM}}(\mathbf{x}_{e,k},\bar{X}_{w}) =
\begin{cases}
1, ~~~\text{if}~\| X_{e,k}- \bar{X}_{w}\|^2_2 \in [\varepsilon_{\mbox{\tiny min}}, \varepsilon_{\mbox{\tiny max}}]\\
0, ~~~~\text{Otherwise.}
\end{cases}
\end{equation}
% The coverage function is dependent on the specific task. In our task, we use the distance between the end effector position and the actual target position to measure the coverage at time $k$.

\textbf{Manipulation temporal condition (MTC)}: In a weed flaming task, the manipulator is required to hold its position for a certain time $\xi$ to ensure sufficient heat transfer. Such an MTC also exists in many other tasks. Recall that $\mathcal{K}_m$ is the time index set for the manipulator to maintain the pose $\mathbf{x}_{e}$.  Let us define MTC 
$\mathds{1}_{\mbox{\tiny MTC}} (\mathbf{x}_{e}, \bar{X}_{w} )$
as a binary function,
\begin{equation}
\label{eq:manipulator_constraint}
    \mathds{1}_{\mbox{\tiny MTC}} (\mathbf{x}_{e},\bar{X}_{w} )=
\begin{cases}
 1, &  \text{if}\ 
~\mathds{1}_{\mbox{\tiny EPCM}}(\mathbf{x}_{e},\bar{X}_{w}) \wedge
\bigl( |\mathcal{K}_m| \ge \xi \bigr) \\ 
0,& \mbox{otherwise,} 
\end{cases}
\end{equation}
where $|\cdot|$ is set cardinality.

%Task constraints are the constraints on the CAPM problem to satisfy in order to perform the task. Actually, it also helps to understand the property of the problem. 

\subsection{Inverse Reachability Region Analysis for Task Constraints\label{sec:inv_reach}}

Based on the inverse kinematics of the arm, the task constraints in Sec.~\ref{ssc:task-constraints} can be used to determine the feasible body states. Denote the joint space of the arm as~$\Theta$. For a given body pose $\mathbf{x}_{b}\in \mathcal{X}_b$ and joint configuration, the forward kinematics function of the arm maps to the last link pose $\mathbf{x}_e \in \mathcal{X}_e$,  $f: \Theta \times \mathcal{X}_b \rightarrow \mathcal{X}_e$ and the inverse kinematics function is  
$ \label{eq:inv_kin}
    f^{-1}: \mathcal{X}_b \times \mathcal{X}_e \rightarrow \Theta.
$
The inverse kinematics function $f^{-1}(\mathbf{x}_b, \mathbf{x}_e)$ for a given end effector $\mathbf{x}_e$ and a body pose $\mathbf{x}_b$ may not necessarily have a solution. We use the term $\exists f^{-1}(\mathbf{x}_b, \mathbf{x}_e)$ to indicate the logical truth that there exists a solution. 

\begin{figure}[hbt!]
    \centering
        \includegraphics[page=2,width=3.2 in]{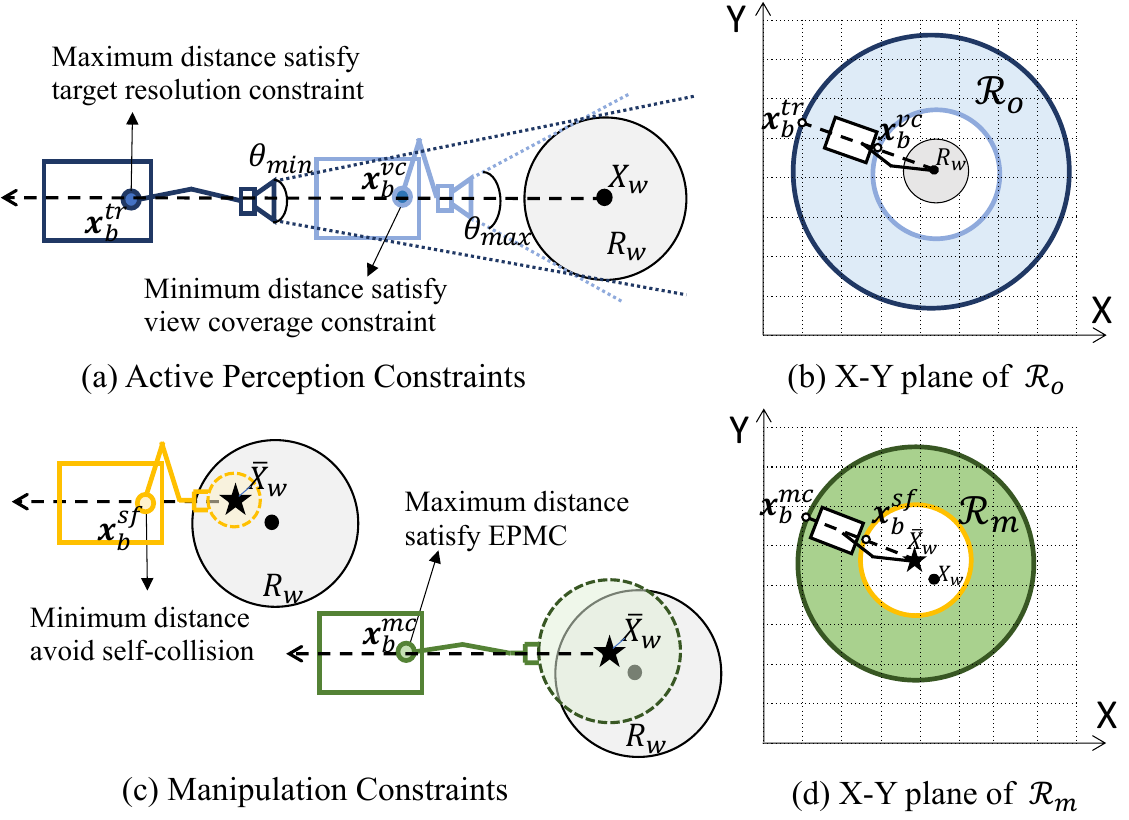}
    \caption{A visualization of feasible body pose sets $\mathcal{R}_o$ and $\mathcal{R}_m$ with color coded boundaries.  }
    \label{fig:regions}
\end{figure}

As shown in Fig.~\ref{fig:regions}(a), for a given TROI $R^{\mbox{\tiny TROI}}_w$, the robot body cannot be too far from it because it needs to satisfy \eqref{eq:target_resolution} with a fixed resolution camera (Assumption a.5). At the farthest reachable point, the dark blue robot arm has to be fully extended to provide good coverage of $R^{\mbox{\tiny TROI}}_w$. Since the robot can approach $R^{\mbox{\tiny TROI}}_w$ from any direction, this boundary condition leads to the dark blue outer circular boundary in Fig.~\ref{fig:regions}(b) as a visualization in the X-Y plane. On the other hand, the robot body cannot be too close to $R^{\mbox{\tiny TROI}}_w$ due to the coverage condition in \eqref{eq:view_coverage}. Being too close cannot maintain a full view of TROI. The boundary condition corresponds to the close-up robot configuration and the inner circular boundary in light blue in Figs.~\ref{fig:regions} (a) and (b), respectively. The region between the two boundaries is the feasible body pose set in the X-Y plane. Define $\mathcal{R}_o(R^{\mbox{\tiny TROI}}_w)$ as the set of feasible body poses, we have
\begin{equation}\label{eq:ro}
\mathcal{R}_o(R^{\mbox{\tiny TROI}}_w) =\{ \mathbf{x}_b |\mathds{1}_{\mbox{\tiny NSV}}(\mathbf{x}_e, R^{\mbox{\tiny TROI}}_w) \wedge \bigl(\exists f^{-1}(\mathbf{x}_b, \mathbf{x}_e)\bigr), \forall \mathbf{x}_{e}  \}, 
\end{equation}
which shows as a blue donut shape co-centered with $R^{\mbox{\tiny TROI}}_w$ in X-Y plane (Fig.~\ref{fig:regions}(b)).

For a given MPOI $\bar{X}_w$, the body pose should be selected such that the arm poses can satisfy \eqref{eq:manipulator_coverage} during the task. Since \eqref{eq:manipulator_coverage} has both minimum and maximum distance thresholds, the set of reachable body poses $\mathcal{R}_m$ also has both inner and outer boundaries color coded in yellow and dark green, respectively, as shown in Figs.~\ref{fig:regions}(c) and (d). Therefore, the shape of $\mathcal{R}_m$ is also donut-like in the X-Y plane.  Mathematically, $\mathcal{R}_m$ is defined as follows, 
\begin{equation}  \label{eq:rm}
    \mathcal{R}_m (\bar{X}_w) :=\{ \mathbf{x}_b | \mathds{1}_{\mbox{\tiny MTC}} (\mathbf{x}_{e},\bar{X}_{w} ) \wedge \bigl(\exists f^{-1}(\mathbf{x}_b, \mathbf{x}_e)\bigr), \forall \mathbf{x}_{e}   \}.
\end{equation}

\subsection{CAPM Problem Types in Mobile Manipulation \label{sec:types}}

\begin{figure}[hbtp!]
    \centering
        \includegraphics[page=3,width=3.5 in]{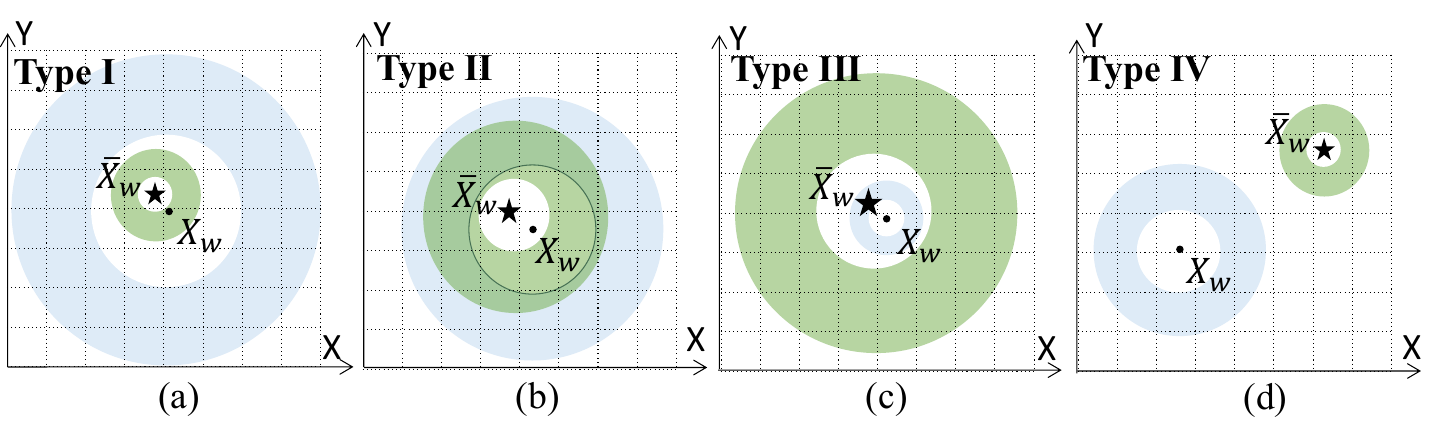}
    \caption{CAPM problem types based on four different $\mathcal{R}_o$ (in lighter blue) and $\mathcal{R}_m$  (in darker green) relationships.}
    \label{fig:prob_cls}
\end{figure}

Defining $\mathcal{R}_o$ and $\mathcal{R}_m$ allows us to classify CAPM problems based on region relationships. Depending on robot sensing and actuation configurations and task requirements, we can classify the CAPM problem into four types, as illustrated in Fig.~\ref{fig:prob_cls}. Before discussing the four types, it is worth noting that $\mathcal{R}_o$ and $\mathcal{R}_m$ are not necessarily co-centered in the X-Y plane, because the former is centered at $X_w$, the center of TROI, and the latter is centered at MPOI $\bar{X}_w$. 

\noindent\textbf{Type I}: 
% Fig.~\ref{fig:prob_cls} (a) shows the type where $\mathcal{R}_o$ is much larger than $\mathcal{R}_m$. Specifically, 
As shown in Fig.~\ref{fig:prob_cls} (a), $\mathcal{R}_o$ is much larger than $\mathcal{R}_m$. In fact, $\mathcal{R}_m$ is completely enclosed by the inner boundary of $\mathcal{R}_o$ and $\mathcal{R}_o \cap \mathcal{R}_m =\emptyset$. This happens when the robot has a strong perception capability that enables the observation for complete TROI at a much further distance. In such problems, active perception and manipulation are decoupled and are essentially a two-step problem that can be solved independently.

\noindent\textbf{Type II }: For Type II (Fig.~\ref{fig:prob_cls} (b)), $\mathcal{R}_o$ has a significant overlap with $\mathcal{R}_m$, but none can fully enclose the other, which means that both $\mathcal{R}_o$ and $\mathcal{R}_m$ are strict subsets of $\mathcal{R}_o\cup \mathcal{R}_m$. 
In such a coupled situation, if a robot can plan its body to be in $\mathcal{R}_o \cap \mathcal{R}_m$, then active perception and manipulation can be addressed simultaneously. 

\noindent\textbf{Type III}: Fig.~\ref{fig:prob_cls} (c) is the type opposite to Type I where $\mathcal{R}_m$ is much larger than $\mathcal{R}_o$. In fact, $\mathcal{R}_o$ is completely enclosed by the inner boundary of $\mathcal{R}_m$ and $\mathcal{R}_o \cap \mathcal{R}_m = \emptyset$. Such a situation occurs when the sensor is very nearsighted, similar to the endoscope camera in minimally invasive surgery~\cite{5625991}. After active perception, the robot needs to retreat for manipulation, which leads to separate solving of active perception and manipulation problem.

\noindent\textbf{Type IV}: Fig.~\ref{fig:prob_cls} (d) shows the case where $\bar{X}_w  \notin R^{\mbox{\tiny TROI}}_w$ and $\mathcal{R}_o \cap \mathcal{R}_m = \emptyset$. This means that the prescan is erroneous or is not up to date. Such a problem usually becomes a search problem like \cite{adu2021probabilistic} instead of a CAPM problem. 

From the analysis, Type II problems fit the nature of mobile manipulator with a common camera sensor, and this is where the CAPM problem actually matters.

\subsection{Energy Cost Function}
In order to solve the CAPM problem while minimizing the energy used, we model the energy cost for the robot moving from a key state $\mathbf{x}_{k_i}$ to another state $\mathbf{x}_{k_{j}}$ as a combination of a fixed initial energy cost for the mobile platform and the variable energy cost. Since we assume that the energy of arm movement is much lower than that of body movement, the arm energy usage is ignored in the cost function. The start energy cost is a fixed cost once movement occurs. We follow the method in \cite{kuffner2004effective} to define the distance metric between two states as $$ d(\mathbf{x}_{k_i},\mathbf{x}_{k_{j}}) = \| X_{b,k_{j}} - X_{b,k_i} \|_2^2 + \beta(1- \| \mathbf{q}_{b,k_{j}} \cdot \mathbf{q}_{b,k_i} \|),$$ where $\beta\ge0$ and determined by specific robot setup.
% Define $X_{b,k_{i+1}}, X_{b,k_i}\in \mathbb{R}^3$ as the 3D position components of the robot poses $\mathbf{x}_{b,k_{i+1}}, \mathbf{x}_{b,k_i} \in SE(3)$, respectively. 
Denote the indicator function for body movement as follow:
\begin{equation}
    \mathds{1}_{\mbox{\tiny b-move}}( \mathbf{x}_{k_{j}} , \mathbf{x}_{k_i} )= \begin{cases}
 1, &  \text{if}~ d(\mathbf{x}_{k_i},\mathbf{x}_{k_{j}})> 0 \\ 
0,& \mbox{otherwise.} 
\end{cases}
%     \left\{\begin{matrix}
%  1, & \| X_{b,k_{i+1}} - X_{b,k_i} \|_2^2 > 0 \\ 
% 0,& \mbox{otherwise} 
% \end{matrix}\right. ,
\end{equation}
% We define fixed initial energy cost for each body motion as,
% \begin{equation}
%     c_s(\mathbf{x}_{k_i},\mathbf{x}_{k_{j}}) =  \mathds{1}_{\mbox{\tiny b-move}}( \mathbf{x}_{b,k_{j}} , \mathbf{x}_{b,k_i} ). \label{eq:cs}
% \end{equation}
% The variable energy cost $c_m$ is the distance of body movement with energy coefficient $\gamma$,
% \begin{equation}
% c_m(\mathbf{x}_{k_i},\mathbf{x}_{k_{j}}) = \gamma d(\mathbf{x}_{k_i},\mathbf{x}_{k_{j}}). \label{eq:cm} 
% \end{equation}
The overall energy cost function is defined by the initial energy cost(first term) and variable energy cost(second term) in the case where body only move once between $\mathbf{x}_{k_i},\mathbf{x}_{k_{j}}$:
\begin{equation}
  c(\mathbf{x}_{k_i},\mathbf{x}_{k_{j}}) = \mathds{1}_{\mbox{\tiny b-move}}( \mathbf{x}_{k_{j}} , \mathbf{x}_{k_i} ) + \gamma d(\mathbf{x}_{k_i},\mathbf{x}_{k_{j}}).
\end{equation}

\subsection{Coupled two-stage key states planning}

With the constraint and energy cost introduced, we are ready to solve the CAPM problem. Intuitively, from the perspective of reducing energy cost, the overall planned key state sequence should contain fewer body movements. The ideal situation is that the robot body only needs to move once to satisfy the requirements of both active perception and manipulation tasks, as shown in the upper branch of Fig.~\ref{fig:keystate}(c). Through the analysis in Sec.~\ref{sec:types}, we know that it is possible for the Type II problem. However, for unknown $\bar{X}_w$, it is unlikely that this can be guaranteed. 
% The problem is probabilistic in nature. 

From Sec.~\ref{ssc:operation-sequence}, there are two key body states $\mathbf{x}_{b,k_1}$ and possible $\mathbf{x}_{b,k_3}$ that determine the energy use. These two become the key decision variable in our planning problem. However, they cannot be obtained in advance because $\bar{X}_w$ is unknown prior to close-up observation. The manipulation constraint cannot be evaluated beforehand. More specifically, $\mathbf{x}_{b,k_1}$ and MPOI jointly determine whether additional body movement is needed to achieve manipulation. Denote the probability that the body state for active perception $\mathbf{x}_{b,k_1}$ is also feasible for manipulation as $p(\mathbf{x}_{b,k_1})$. Since we know $\bar{X}_w \in R^{\mbox{\tiny TROI}}_w$, we can establish a probability distribution for the estimate of the target position
$ \hat{X}_w \sim \mathcal{N}(X_w, \Sigma_w)$ 
using the target region of interest described by the center $X_w$ and the radius $r_w$. The probability $p(\mathbf{x}_{b,k_1})$ can be derived as
\begin{align}
     p(\mathbf{x}_{b,k_1}) &= p(\mathbf{x}_{b,k_1} \in R_m(\hat{X}_w)|\hat{X}_w) \\ &= \int p(\hat{X}_w) \mathds{1}(\mathbf{x}_b,\hat{X}_w)  d \hat{X}_w,
\end{align}
where $\mathds{1}(\mathbf{x}_b,\hat{X}_w)  =   \begin{cases}
 1, &  \text{if}~ \mathbf{x}_{b,k_1} \in R_m(\hat{X}_w) \\ 
0,& \mbox{otherwise.} 
\end{cases}
$

Except for $\mathbf{x}_{b,k_1}$, other states are deterministic given $\mathbf{x}_{b,k_1}$ and $\bar{X}_w$. The minimum energy CAPM problem formulation in \eqref{eq:min-exp-energy} is reduced to the following:
\begin{align}
   \min_{\mathbf{x}_{b,k_1}, \mathbf{x}^l_{b,k_3}} \quad &  c_0 + p(\mathbf{x}_{b,k_1})c_u + (1-p(\mathbf{x}_{b,k_1})) c_l \label{eq:min_2stage}\\
\nonumber\textrm{s.t.}  \quad &  \mathbf{x}_{b,k_1} \in \mathcal{R}_o(R^{\mbox{\tiny TROI}}_w) ~~\text{and}~~ \mathbf{x}^l_{b,k_3} \in \mathcal{R}_m(X_w), 
\end{align}
where the energy cost $c_0 = c(\mathbf{x}_{b,0},\mathbf{x}_{b,k_1})$, $ c_u = c(\mathbf{x}_{b,k_1},\mathbf{x}_{b,k_{\mbox{\tiny max}}}) $, and $ c_l =c(\mathbf{x}_{b,k_1},\mathbf{x}^l_{b,k_3})+ c(\mathbf{x}^l_{b,k_3},\mathbf{x}_{b,k_{\mbox{\tiny max}}}) $.

In addition, the state $\mathbf{x}_{b,k_{\mbox{\tiny max}}}$ is the final state in which the robot body should arrive. This state can be determined by the current target and the next target's TROI, e.g. the middle point between two target's center. If the current target is the last target, $\mathbf{x}_{b,k_{\mbox{\tiny max}}}$ is the predefined end pose given by the user. 
By solving the above optimization, we obtain the key states for the robot base. Also, finding the best manipulator pose for a given base pose and intermediate states beyond key states is trivial, and we omit the details here. A viable method is proposed in \cite{arm_energy}. 
After the robot executes the key states $\mathbf{x}_{b,k_1}$ and $\mathbf{x}_{e,k_2}$ to perform active peception, the precise MPOI $\bar{X}_w$ can be perceived. If the planned location for the next state $\mathbf{x}^l_{b,k_3}$ does not satisfy the EPMC constraint, re-planning is needed for finding $\mathbf{x}^l_{b,k_3}$. The re-planning objective is to minimize $c_l$ subject to $\mathbf{x}^l_{b,k_3} \in \mathcal{R}_m(\bar{X}_w)$.

% The navigation cost for A, C, E in the Fig.~\ref{fig:pipeline} and Fig.~\ref{fig:illustration} can be represented as $c_A = c(\mathbf{x}_{b,0},\mathbf{x}_{b,k_1})$, $c_E = c(\mathbf{x}_{b,k_1},\mathbf{x}_{b,k_{\mbox{\tiny max}}})$, and in the case where manipulation is required base to move $c_{CE} = c(\mathbf{x}_{b,k_1},\mathbf{x}^l_{b,k_3})+ c(\mathbf{x}^l_{b,k_3},\mathbf{x}_{b,k_{\mbox{\tiny max}}})$.

\begin{comment}
    
The optimal key states of the arm with the minimum energy consumption are determined by the observation result $\hat{X}_w$ and the result of the optimal key state of the base state of the robot. We can use the existing method to find the state \cite{arm_energy}. Specifically, once the robot finishes executing the key states for active perception and collect image from the camera, the MPOI $\bar{X}_w$ is provided by the detection algorithm based on \cite{Xie_2021} using the camera image. Then, the key states for manipulation, which depend on the condition \eqref{eq:base_position}, can be solved using \cite{arm_energy}.

% Specifically, the optimal manipulator pose for observation is 

\begin{figure}[htbp!]
    \centering
    \includegraphics[page=2,width=3.0in]{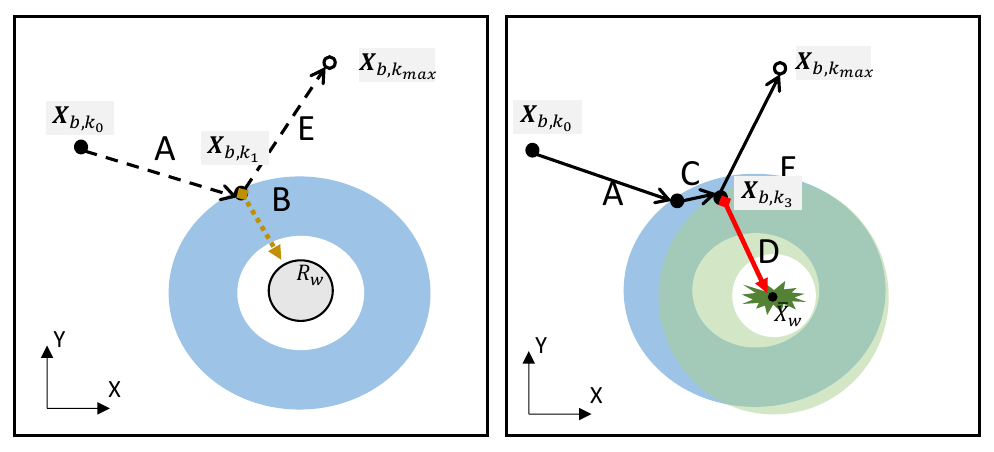}
    \caption{Algorithm Illustration. The characters on the edge correspond to pipeline Fig.~\ref{fig:pipeline}. }
    \label{fig:illustration}
\end{figure}
\end{comment}

\section{Experiments}
In the experiment, the mobile manipulator is a Boston Dynamic Spot Mini\texttrademark\ with a Unitree Z1\texttrademark\  arm. We use IKFast \cite{diankov_thesis} to compute the inverse kinematics for the task constraint regions $\mathcal{R}_o$ and $\mathcal{R}_m$. The experiment includes two parts: the simulation for the comparison of algorithm performance with naive algorithms and physical experiments.
% for  are performed to validate the algorithm in the precision agriculture application for weed flaming. 
\subsection{Simulation}
In simulation, we compare the naive planners (Alg. a) and our proposed method (Alg. c). Also, we consider the variant of our algorithm, the decoupled version (Alg. b). All algorithms minimize the expected energy cost given start state $\mathbf{x}_{k_0}$ and end state $\mathbf{x}_{k_{\mbox{\tiny max}}}$.
\begin{itemize}
\itemsep0em
    \item[a.] Deterministic planner. This planner considers the geometric center of TROI $X_w$ as the MPOI, that is, assume $\bar{X}_w = X_w$ and solve $\min_{\mathbf{x}_{b,k_1}} c(\mathbf{x}_{b,k_0},\mathbf{x}_{b,k_1})+c(\mathbf{x}_{b,k_1},\mathbf{x}_{b,k_{\mbox{\tiny max}}})$, subject to $\mathbf{x}_{b,k_1} \in \mathcal{R}_m(X_w)$.
    % Specifically, this planner searches for the shortest path from $\mathbf{x}_{k_0}$ to $\mathbf{x}_{k_{\mbox{\tiny max}}}$ with one stop that satisfies the manipulation constraint. 
    % \item[b.] Manipulator planner with passive observation. \textcolor{blue}{(I'm considering design this planner adapt from \cite{adu2021probabilistic,garrett2020online}. In these methods, the robot/target state modeling is probabilistic, but the modeling is for the general task and motion planning. It will be difficult to compare directly with either of these methods. )}
    \item[b.] Decoupled active perception and manipulation planner. This planner has an action sequence of the lower branch in Fig.~\ref{fig:keystate}(c), without considering the possibility of the upper branch in the figure when solving the key states.
    \item[c.] CAPM Planner that solves \eqref{eq:min_2stage}. 
\end{itemize}

Other experiment parameters are chosen on the basis of typical field conditions. We consider the operating height of Spot Mini is fixed at $0.8$m. We randomly generate $N=1000$ different TROI $X_w= (x_w,y_w,0)$ within the $3\times3$m$^2$ workspace with radius $r_w \in [0.2\text{m},0.3\text{m}]$. The MPOI is sampled through $ \bar{X}_w \sim \mathcal{N}(X_w, \Sigma_w)$ where $\Sigma_w = r_w \mathbf{I}_{2\times2}$. The energy-cost coefficient $\gamma = 2$. To simulate the scenario with different weed density, for each sample TROI we generate 5 different path lengths.

\begin{comment}
\begin{figure}[htbp]
\centering
\subfigure[\label{fig:exp_two_step}]{\includegraphics[width=1.7in]{Image/output_3.png}}
\hspace*{-.15in}\subfigure[\label{fig:exp_two_step}] {\includegraphics[width=1.7in]{Image/output_5.png}}
\caption{ Blue double ring and green double rings are the result of $\mathbf{R}_o$ and $\mathbf{R}_m$ solved by the inverse kinematic solver using our platform. Given the start and end base positions (red dots), there are two possible types of path given by CAPM. (a) and (b) correspond to the upper and lower branches in Fig.~\ref{fig:keystate}.  }
\label{fig:exp_config_res}
\end{figure}
\end{comment}

To compare the baseline and variants of our algorithm, we employ two performance metrics: success rate ($\%$) and average energy cost ($\mbox{Avg}(c)$). The success of the task is defined as the pose of the manipulator that satisfies the EPMC constraint \eqref{eq:manipulator_coverage} with $\varepsilon_{\mbox{\tiny min}} =0.05$ and $\varepsilon_{\mbox{\tiny max}}=0.10$. The success rate is the ratio between the total number of successful algorithms and $5N$. The result is listed in Table~\ref{tab:sim_res}. It is clear that although the baseline algorithm (Alg. a) achieves a lower energy cost without considering the active perception task, its success rate is the lowest due to the lack of MPOI information. The decoupled active perception and manipulation planner (Alg.b) successfully executes all tasks, but the energy cost is higher than that of our proposed CAPM planner (Alg.c).
To compare the energy savings of the coupled version algorithm with the decoupled version, we define the energy saving coefficient as $\eta_{bc} = (\mbox{Avg}(c)_b -\mbox{Avg}(c)_c)/ \mbox{Avg}(c)_c $ where subscripts $a,b,$ and $c$ refer to Alg.a, Alg.b, and Alg.c, respectively. It is clear that the CAPM algorithm works better when the distance traveled between targets is short (i.e., dense weed distribution). This is desirable.
\begin{table}[htbp]
	\centering
    \caption{Simulation results of the 3-algorithm comparison}
	\label{tab:sim_res}
	%\footnotesize{
    %\scalebox{1.2}{
	\begin{tabular}{c | c |c c c c c  }
		\toprule[0.8pt]
		%\hline
		%\midrule
                & & \multicolumn{5}{ |c }{\mbox{Avg}($c$) at different path length (m)} \\ \cline{3-7}
        Alg. & $\%$  & 2.75 & 3.25 & 3.75 & 4.25 & 4.75 \\ \hline
a &  38 &  3.29   &  3.56 & 4.01 & 4.21  & 4.52  \\
b  & 100 &  4.35 & 4.62 & 4.89 & 5.18  & 5.48  \\
c  & 100 & 3.82 &4.20 & 4.46  & 4.73  & 5.09  \\
   \hline
\multicolumn{2}{c|}{$\eta_{bc}$}& \textbf{0.14}	& 0.10	&0.10	&0.09	&0.08   \\
\hline
\end{tabular}
\centering
\vspace{-4mm}
\end{table}

\subsection{Physical experiment}
We evaluate our system on the physical platform. The CAPM planner is able to efficiently and precisely burn down the weeds. More details are given in the video attachment.

\section{Conclusion and Future Work}

We presented a CAPM problem where a mobile manipulator must obtain a close-up view before performing its manipulation task. Due to the fact that the perception results determine the follow-up manipulation, the proposed task planning approach can exploit the overlapping observation and manipulation reachable sets to reduce the overall expected energy usage while guaranteeing the task success rate. We implemented our CAPM algorithm and tested it in both simulation and physical experiments. The results confirmed our design and showed that our algorithm has a lower energy consumption compared to a typical two-stage decoupled approach while still maintaining a success rate for the task 100\%. In the future, we will improve the proposed algorithm to handle multiple TROI and multiple mobile manipulator problems.    

\section*{Acknowledgement}
{\small The authors would like to thank Dylan Shell, Jason O'Kane, Fengzhi Guo, Aaron Kingery, Yingtao Jiang, and Hojun Ji for their inputs and feedbacks.}

\bibliographystyle{IEEEtran}
\bibliography{syxie}

\end{document}